\documentclass[letterpaper, 10 pt, conference]{ieeeconf}  

\IEEEoverridecommandlockouts                              

\usepackage{amsmath}
\usepackage{nccmath}
\usepackage{algorithmic}
\usepackage{algorithm}
\usepackage{amsfonts}
\usepackage{caption}
\usepackage{amsmath}
\usepackage{multirow}
\usepackage{multicol}
\usepackage{caption}
\usepackage{subcaption}
\usepackage{graphicx}
\usepackage{verbatim}
\usepackage{amsmath,amssymb}
\usepackage{mathtools}
\usepackage{amsmath,mleftright}
\usepackage{xparse}
\usepackage{tikz}
\usepackage{graphicx}
\usetikzlibrary{positioning}

\DeclareMathOperator*{\argmax}{arg\,max}
\DeclareMathOperator*{\argmin}{arg\,min}
\newcommand*{\argmaxl}{\argmax\limits}

\NewDocumentCommand{\evalat}{sO{\big}mm}{%
  \IfBooleanTF{#1}
   {\mleft. #3 \mright|_{#4}}
   {#3#2|_{#4}}%
}

\title{\LARGE \bf
Visual Measurement Integrity Monitoring for UAV Localization}

\author{Chengyao Li$^{1}$ and Steven L. Waslander$^{2}$
\thanks{*This work was supported by the NSERC Canadian Robotics Network.}
\thanks{$^{1}$C. Li, and S. L. Waslander are with the Toronto Robotics and AI Laboratory (TRAILab) in the Institute for Aerospace Studies, University of Toronto.
        {email: (chengyao.li@mail.utoronto.ca)}, {(stevenw@utias.utoronto.ca)}}%
}

\begin{document}

\maketitle
\thispagestyle{empty}
\pagestyle{empty}

\begin{abstract}

   Unmanned aerial vehicles (UAVs) have increasingly been adopted for safety, security, and rescue missions, for which they need precise and reliable pose estimates relative to their environment. To ensure mission safety when relying on visual perception, it is essential to have an approach to assess the integrity of the visual localization solution. However, to the best of our knowledge, such an approach does not exist for optimization-based visual localization. Receiver autonomous integrity monitoring (RAIM) has been widely used in global navigation satellite systems (GNSS) applications such as automated aircraft landing. In this paper, we propose a novel approach inspired by RAIM to monitor the integrity of optimization-based visual localization and calculate the protection level of a state estimate, i.e. the largest possible translational error in each direction. We also propose a metric that quantitatively evaluates the performance of the error bounds. Finally, we validate the protection level using the EuRoC dataset and demonstrate that the proposed protection level provides a significantly more reliable bound than the commonly used $3\sigma$ method. 

\end{abstract}

\section{INTRODUCTION}

In recent years, unmanned aerial vehicles (UAVs) have found success in challenging applications such as event safety, search and rescue, and security surveillance. For some specific operating domains such as indoors, underground or in urban centers, global navigation satellite systems (GNSS) information is not always available or reliable onboard the UAV. Hence, the feasibility of UAV deployment in these locales depends highly on the accuracy of the vision-based localization as a complement and replacement to GNSS systems. Failure to correctly perform the localization task with the sensors available to the UAV might lead to serious damage to the vehicle or to people in the vicinity. It is therefore imperative to not only perform accurate UAV localization with visual information, but to understand the reliability of the current localization estimate and to minimize the possibility of undetected failures in the visual localization process.  

Many UAV visual localization or visual simultaneous localization and mapping (SLAM) algorithms have been developed that can provide robust state estimation over an extended period of time, such as~\cite{mur2017orb, leutenegger2013keyframe}. Although these algorithms are able to operate with a low failure rate, it remains possible to further improve the robustness of UAV state estimation by continuously assessing the integrity of the state estimate computed onboard the UAV. Integrity measures the degree of trust that can be placed on the correctness of the localization solution~\cite{ochieng2002assessment}.

In robot visual localization, 3$\sigma$ ($\pm$3 standard deviations) is often used as a measure of the uncertainty bounds on state estimates, which corresponds to a 99.7\% probability that the ground truth state is within the region. However, 3$\sigma$ only bounds the error of the state assuming there are no outliers in the measurements, that is, no measurements are drawn from outside the modeled measurement distribution. For real-life UAV visual localization, it is unreasonable to assume that the visual front-end works perfectly and there are never any outliers in the measurements at any time, as visual front-ends rely on correspondence of features from frame to frame, a process that is susceptible to error. As a result, the 3$\sigma$ approach is often too aggressive and does not bound the error of the true state of the UAV.

The work of this paper is inspired by received autonomous integrity monitoring (RAIM) which is an integrity monitoring algorithm developed to assess the integrity of the GNSS signals. It is especially used in automated aircraft landing and other safety-critical GNSS applications. RAIM first uses redundant measurements to check the consistency of the measurements received, and detects possible faulty measurements from individual satellites. Afterwards, it determines a Horizontal Protection Level (HPL) and a Vertical Protection Level (VPL), which are the maximum errors in horizontal and vertical directions that the fault detection algorithm is not expected to detect~\cite{walter1995weighted}. The idea of adapting RAIM for robot localization has been recently developed for filter-based algorithms~\cite{2019imekf, 2019imekf2}, but not for optimization-based algorithms. 

Although integrity monitoring is a well-developed area for GNSS applications, it is still non-trivial to adapt it in optimization-based UAV visual localization. Firstly, in most of the UAV visual localization algorithms, the system relies on features to provide measurements instead of satellites, and there are considerably more measurements compared with GNSS applications. Secondly, in RAIM, it is rare for two satellites to be faulty at the same time. However, for visual localization, the absolute number of incorrect measurements, or outliers, is much higher, and the outlier ratio can often be greater than 10\%. As a result, a different outlier rejection method is needed to handle multiple outliers. In addition, in GNSS applications, each satellite provides one measurement, but in visual localization, each feature gives 2 or 3 measurements for the monocular and rectified stereo cases, respectively.

In this paper, we propose a novel approach to monitor the integrity of the state estimation in optimization-based UAV visual localization. We first detect and isolate inconsistent visual measurements using a statistical method that is based on~\cite{tong2011batch} and expands on our previous work~\cite{das2014outlier}. We then calculate the largest translational error that can exist in the state of the UAV. The main contributions of this paper are:

\begin{itemize}
    \item A novel metric called relaxed bound tightness that quantitatively evaluates the performance of error bounds for applications where the error is assumed to follow a Gaussian distribution.
    \item An approach to determine an approximate upper bound of the error in the state estimation of the UAV in optimization-based visual localization that is significantly more reliable than the typical 3$\sigma$ approach.
\end{itemize}

The remainder of this paper is outlined as follows: related work is discussed in Sec.~\ref{related_work}, details of the outlier rejection and protection calculation algorithms are discussed in Sec.~\ref{algorithms}, the proposed metric for evaluating error bounds is shown in Sec.~\ref{metric}, and finally, validation results of the proposed approach are shown in Sec.~\ref{results}.

\section{RELATED WORK}\label{related_work}
\subsection{Receiver Autonomous Integrity Monitoring}
The concept of integrity monitoring has been developed and applied to GNSS applications. It is a necessary component for safety-critical applications where unreliable solutions might lead to serious injury or death. RAIM is one of the more commonly used integrity monitoring algorithms for aviation applications. It has two basic functions. The first is to detect whether there is a satellite failure using fault detection and exclusion algorithm. The second is to calculate the HPL and VPL, which are the largest horizontal and vertical errors that the fault detection algorithm is not expected to detect~\cite{liu2005gps}. Brown et al.~\cite{brown1994gps} propose the first mathematically rigorous RAIM algorithm, which uses statistical theory to check the consistency of the redundant GPS measurements and calculates the HPL with the measurement group that provides adequate consistency. Walter et al.~\cite{walter1995weighted} use Brown's method as a basis to develop weighted RAIM. The main idea of their work is that the measurements coming from each individual satellite have different levels of noise. As a result, instead of assuming a fixed covariance value for all the satellites, they propose to trust the measurements from satellites differently by assigning noise with weighted covariance to the measurement models of different satellites. Both methods assume that there is at most one faulty measurement in the system after the fault detection algorithm, and then they calculate the protection levels based on this assumption. Angus~\cite{angus2006raim} revise the algorithms proposed by Brown and Walter to take account of multiple faulty satellites in the measurements. The ability to handle multiple faulty measurements is useful in the context of visual localization because each feature corresponds to 2 or 3 measurements depending on the camera settings. As a result, to take account of 1 faulty feature in visual localization, we need to take account of 2 or 3 faulty measurements.

\subsection{Vision-Based Localization}
Vision-based localization or SLAM algorithms can be divided into two main categories, direct method or feature-based method. Direct methods process the entire image as the measurement and aim to minimize the photometric error~\cite{engel2014lsd}. Feature-based methods extract keypoints from the image and the objective is to minimize reprojection error. ORB-SLAM2~\cite{mur2017orb} is a state-of-the-art feature-based SLAM system for monocular, stereo, and RGB-D cameras. Its SLAM mode performs bundle adjustment to construct the map. ORB-SLAM2 also has a localization mode that allows the reuse of the map to determine the state of the agent by pose optimization. In addition, combining visual and inertial measurements also shows improvement in the performance especially for UAV applications. such as~\cite{leutenegger2013keyframe}. Although these algorithms show promising performance in experiments, it is unreasonable to assume that the measurements are fault-free at all times because finding correspondences of features from frame to frame is a process that is susceptible to error especially in a dynamic environment. The work of this paper is mainly applicable to feature-based visual localization method, and we use the localization mode of ORB-SLAM2 to validate our method, which is shown in Sec.~\ref{results}. 

\subsection{Outlier Rejection For Visual Measurements}
RANSAC~\cite{fischler1981random} has been widely used in vision-based robot localization~\cite{kitt2010visual} to reject feature outliers. The basic idea of RANSAC is to use random sets of samples to form hypotheses and use the other samples to verify these hypotheses, and the hypothesis with the highest consensus is selected to be the inlier set~\cite{scaramuzza20111}. An alternative approach is to use statistical tests to check if the measurement set fits the assumed statistical model, such as Parity Space Approach~\cite{das2014outlier}, and Normalized Innovation Squared (NIS)~\cite{bar2004estimation}. These two methods both assume the noise of the measurement model follows a Gaussian distribution and check whether the weighted sum of squares residual follows a $\chi^2$ distribution. Parity Space Approach does not require the state of the system and it checks the consistency of a group of redundancy measurements by projecting them into the parity space~\cite{das2014outlier}. On the other hand, NIS assumes that an estimated state is known, and it is usually used to check if individual measurement is an outlier. Tong et al.~\cite{tong2011batch} propose a batch innovation test that extends NIS to remove outliers for a batch of measurements. Recently, Tzoumas et al.~\cite{tzoumas2019outlier} propose an adaptive trimming algorithm that also removes outliers based on the value of residuals. Instead of using a fixed threshold, the threshold in this algorithm updates for each iteration. In this paper, we adapt Tong's method and use it with the Parity Space Approach to remove multiple outliers iteratively, which leads to the calculation of protection level.

\section{INTEGRITY MONITORING FRAMEWORK}\label{algorithms}
\subsection{Problem Formulation}
In this section, we show the problem formulation for stereo camera settings. We formulate the vision-based localization as a nonlinear pose optimization. The objective is to determine the camera state at the current frame by minimizing the reprojection error of features. We follow the notation introduced by Barfoot~\cite{barfoot2017state}. There are two frames, the inertial frame $\mathcal{F}_i$, and the camera frame $\mathcal{F}_c$ which corresponds to the left camera center. The state of the camera is defined as $\textbf{x} = $\{${\textbf{r}^{ci}_i}$, $\textbf{C}_{ci}$\}, which is the transformation from the inertial frame to the camera frame, where ${\textbf{r}^{ci}_i} \in \mathbb{R}^3$ and $\textbf{C}_{ci} \in \mathbb{SO}(3)$. Before performing localization, a map is built with features, and $\textbf{r}^{p_ji}_i$ denotes the position of the feature $j$ in the inertial frame. The feature position is transformed from the inertial frame to the camera frame first and then projected to the image coordinate plane using the following equations:

\begin{equation}
\textbf{r}^{p_j c}_{c} =\begin{bmatrix}x\\ y\\z\end{bmatrix} =\textbf{C}_{ci}(\textbf{r}^{p_ji}_i-\textbf{r}^{ci}_i)
\end{equation}

\begin{align}
\begin{bmatrix} u_l\\ v_l\\ \delta_d\end{bmatrix} = \pi(\textbf{r}^{p_j c}_{c}) =
\frac{1}{z}
\begin{bmatrix} f_ux \\ f_vy \\ f_ub\end{bmatrix} + 
\begin{bmatrix} c_u\\ c_v \\ 0 \end{bmatrix} + \mathbf{e}_j
\end{align}
\begin{align}
\label{measurement noise}
        \mathbf{e}_j \sim \mathcal{N}(\mathbf{0},\;\mathbf{Q}_j)
\end{align}
where $u_l,v_l$ are the left image coordinates, $\delta_d$ is the disparity, $f_u, f_v$ are the focal length, $c_u, c_v$ are the principle points, b is the baseline, $\pi$ is the stereo camera projection function, and $\mathbf{e}_j$ is the measurement noise that is assumed to follow a Gaussian distribution with covariance matrix $\mathbf{Q}_j$. At each frame, the pose optimization can be formulated as follows:
 
\begin{align}
       \label{eq:optimization}
\mathbf{x^*} =\Bigl\{{\textbf{r}^{ci*}_i,\textbf{C}^*_{ci}} \Bigr\} = \argmin_{\mathbf{r}_i^{ci},\mathbf{C}_{ci}} 
&\sum_{j}^{}  \rho(\mathbf{e}_{y,j}^T\mathbf{Q}^{-1}_j\mathbf{e}_{y,j})
\end{align}
where $\rho$ is the Huber loss function, and $\mathbf{e}_{y,j}$ is the measurement error term for feature $j$ and can be determined using the formula below:

\begin{equation}
\mathbf{e}_{y,j}(\mathbf{x}) = \mathbf{y}_{j} - \pi(\textbf{C}_{ci}(\textbf{r}^{p_ji}_i-\textbf{r}^{ci}_i))
\end{equation}
where $\mathbf{y}_{j}$ is the measurement for feature $j$.

\subsection{Fault Detection and Exclusion}
This subsection shows how to apply fault detection and exclusion to visual measurements in a manner inspired by RAIM~\cite{walter1995weighted} using Parity Space Method. First, we have the measurement model for each feature $j$:

\begin{ceqn}
    \begin{align}
            \mathbf{y}_{j} = h_j(\textbf{x}) + \mathbf{e}_j
    \end{align}
\end{ceqn}
where the measurement function for feature $j$ is $h_j(\textbf{x}): \mathbb{R}^6 \Rightarrow \mathbb{R}^{3}$, defined as 
\begin{ceqn}
    \begin{align}
             h_j(\textbf{x})  = \pi(\textbf{C}_{ci}(\textbf{r}^{p_ji}_i-\textbf{r}^{ci}_i)).
    \end{align}
\end{ceqn}
Then we linearize the measurement function $h_j(\textbf{x})$ about an operating point $\textbf{x}_0$ to obtain the linearized measurement model:
\begin{ceqn} 
    \begin{align}      
    \label{eq:linearized_measurement}
            d\mathbf{y}_{j} = \mathbf{H}_j\cdot d\textbf{x} + \mathbf{e}_j
    \end{align}
\end{ceqn}
where $d\textbf{x} \in \mathbb{R}^6$ is the state perturbation, $\mathbf{H} \in \mathbb{R}^{3\times6}$ is the Jacobian of the measurement model, and $d\textbf{y}_j \in \mathbb{R}^3$ is the shifted measurements according to the operating point $\mathbf{x}_0$ for feature $j$. We make the following definitions by stacking $d\mathbf{y}_j$, $\mathbf{H}_j$, and covariance matrix $\mathbf{Q}_j$ for $N$ features observed in the current frame:    
\begin{ceqn}
    \begin{align}
            d\mathbf{y} = \begin{bmatrix} d\mathbf{y}_1\\\vdots \\  d\mathbf{y}_N \end{bmatrix}, 
            \mathbf{Q}  = \begin{bmatrix}  \mathbf{Q}_1 & &\\& \ddots & \\  & & \mathbf{Q}_N  \end{bmatrix}, 
            \mathbf{H} = \begin{bmatrix} \mathbf{H}_1\\\vdots \\  \mathbf{H}_N \end{bmatrix}.
    \end{align}
\end{ceqn}
The stacked version of Eq.~\ref{eq:linearized_measurement} is:
\begin{ceqn} 
    \begin{align}   \label{eq:stacked linearized measurement model}   
            d\mathbf{y} = \mathbf{H}\cdot d\textbf{x} + \mathbf{e}
    \end{align}
\end{ceqn}
where
\begin{align}
        \mathbf{e} \sim \mathcal{N}(\mathbf{0},\;\mathbf{Q}).
\end{align}
We define information matrix $\mathbf{W}$ which is the inverse of $\mathbf{Q}$:
\begin{ceqn}
    \begin{align}
\textbf{W} =\mathbf{Q}^{-1}
    \end{align}
\end{ceqn}
The estimated perturbation is the weighted least squares solution of Eq.~\ref{eq:stacked linearized measurement model}, which is:
\begin{ceqn}
    \begin{align}\label{sol_linear}
        d\hat{\mathbf{x}} = (\mathbf{H}^T\mathbf{W}\mathbf{H})^{-1}\mathbf{H}^T\mathbf{W}d\mathbf{y}.
    \end{align}
\end{ceqn}
The residual can be further calculated as follows:
\begin{ceqn}
    \begin{align} \label{eq:residual}
       \mathbf{ \epsilon} = d\mathbf{y} - \mathbf{H}d\hat{\mathbf{x}} = (\mathbf{I} - \mathbf{H}(\mathbf{H}^T\mathbf{W}\mathbf{H})^{-1}\mathbf{H}^T\mathbf{W})d\mathbf{y}.
    \end{align}
\end{ceqn}	
However, if we have outliers in the measurements, Eq.\ref{eq:stacked linearized measurement model} is rewritten as: 
\begin{ceqn}
    \begin{align} \label{eq:fault}
            d\mathbf{y} = \mathbf{H}\cdot d\textbf{x} + \mathbf{e} + \mathbf{f}
    \end{align}
\end{ceqn}
where $\mathbf{f} \in \mathbb{R}^{3N}$ is the fault vector which models the error in the measurements under the assumption that there are outliers in the measurements~\cite{das2014outlier}. Eq.~\ref{eq:residual} has to be rewritten to be:
\begin{ceqn}
    \begin{align} 
       \mathbf{ \epsilon} = (\mathbf{I} - \mathbf{H}(\mathbf{H}^T\mathbf{W}\mathbf{H})^{-1}\mathbf{H}^T\mathbf{W})(d\mathbf{y} -\mathbf{f})
    \end{align}
\end{ceqn}	
We then calculate the weighted sum of squares residual:
\begin{ceqn}
    \begin{align}
        \lambda = \mathbf{\epsilon}^T\mathbf{ W} \mathbf{\epsilon}.
    \end{align}
\end{ceqn}
We can use $\lambda$ to determine whether outliers exist in the measurement set and $\sqrt{\lambda}$ is defined as the test statistic. If there are no outliers in the measurement data, $\lambda$ follows a central $\chi^2$ distribution with $n-m$ degrees of freedom based on the Parity Space Method~\cite{liu2005gps}. Here, $n$ is the number of measurements and $m$ is the number of states, and in the case of vision-based localization $n = 3N$ and $m = 6$. If there are outliers in the measurements, $\lambda$ follows a non-central  $\chi^2$ distribution. The weighted sum of squares  residual, $\lambda$, can be used to determine whether the position solution is outlier-free. A false alarm threshold $\delta$ can be selected analytically based on a desired probability of false alarm $P_{fa}$, by computing the $1-P_{fa}$ quantile of the central $\chi^2$ distribution with $3N-6$ degrees of freedom. If $\lambda < \delta$, the test indicates that the measurements are consistent with each other and it is hypothesized that there is no outlier in the measurement set, and if $\lambda > \delta$, there is a $1-P_{fa}$ probability that there are one or more outliers in the measurement set.
    
If the measurement set passes the outlier detection algorithm, we can directly calculate the protection level in~\ref{PL}. If it is found that the measurement set contains outliers, we need to perform outlier rejection. Unlike GNSS applications where it is rare for two satellites to fail at the same time, the outlier ratio for visual measurements is significantly higher. As a result, a different approach is required to remove the outliers in the measurements for vision-based localization. 
    
We modify the approach in~\cite{tong2011batch} and propose the Iterative Parity Space Outlier Rejection (IPSOR) method to remove multiple outliers for integrity monitoring. Both algorithms calculate the weighted sum of squares residual on a group of measurements and remove outliers iteratively until the remaining measurements pass the residual threshold test. Specifically, the IPSOR strategy removes the feature that contributes the most to the test statistic iteratively and classifies them as outliers until the test statistic $\lambda$ is less than the threshold $\delta$ given the number of measurements. Afterward, we classify the remaining measurements as inliers and calculate the test statistic again. If the updated test statistic is less than the updated threshold $\delta$, the outlier rejection is completed. Otherwise, we repeat this process. If the number of inliers is less than a threshold in this process, we deem that there are too many outliers in the measurements and the localization position is unsafe. This method relies on the assumption that inlier measurements agree with each other and outlier measurements vary much more widely. Using the Gauss-Newton algorithm, the IPSOR method can be applied iteratively to obtain a better linearization state $\mathbf{x}_0$ and avoid removing inliers.
    
\begin{algorithm}
	\caption{Given a set of $N$ 3D features $Y$ and an initial state estimate $\mathbf{x}_0$, we can obtain a set of inliers $Y_{inlier}$ which pass the outlier detection algorithm using IPSOR method.}
	\begin{algorithmic}[1]
	\STATE{$Y_{inlier} \leftarrow Y$}
	\STATE{$\mathbf{H} \leftarrow [\frac{\partial}{\partial x} h({\mathbf{x}})_1 \vert_{{\mathbf{x_0}}}, \frac{\partial}{\partial x} h({\mathbf{x}})_2 \vert_{{\mathbf{x_0}}} \ldots \frac{\partial}{\partial x} h({\mathbf{x}})_N \vert_{{\mathbf{x_0}}}]$}
	\STATE{$d\mathbf{y} \leftarrow [ d\mathbf{y}_1 \; d\mathbf{y}_2 \ldots   d\mathbf{y}_{N}]$}
	\STATE{$d\hat{\mathbf{x}} = (\mathbf{H}^T\mathbf{W}\mathbf{H})^{-1}\mathbf{H}^T\mathbf{W}d\mathbf{y} $}
	\STATE{$       \mathbf{ \epsilon} = d\mathbf{y} - \mathbf{H}d\hat{\mathbf{x}} = (\mathbf{I} - \mathbf{H}(\mathbf{H}^T\mathbf{W}\mathbf{H})^{-1}\mathbf{H}^T\mathbf{W})d\mathbf{y} $}
	\STATE{$\lambda \leftarrow \epsilon^T\mathbf{W}\epsilon$}
	\STATE{$\delta \leftarrow \chi^2_{(3N-6,1-P_{fa})}$}

	\WHILE{$ \lambda >  \delta$ }
	\FOR{$j < N$}
    	\STATE{$ \mathbf{\lambda}_i(j)\leftarrow \epsilon_j^TW(j,j)\epsilon_j$}
    	\STATE{$j \leftarrow j + 1$}
    \ENDFOR
		\WHILE{$ \lambda > \delta$}
				\STATE{$ \lambda_{max} \leftarrow \max(\mathbf{\lambda}_i(j))$}
				\STATE{$ i^* \leftarrow \argmaxl_{i}(\mathbf{\lambda}_i(j))$}		
				\STATE{$ \lambda \leftarrow \lambda - \lambda_{max}$}
			    \STATE{$N \leftarrow N -1$}
			    \STATE{$Y_{inlier} \leftarrow Y_{inlier}  - Y_{i^*}  $}
	            \STATE{$\delta \leftarrow \chi^2_{(3N-6,1-P_{fa})}$}
		\ENDWHILE
	\IF{$N_I < N_T$}
	    \STATE{\emph{Break}}
	\ENDIF
	\STATE{$\mathbf{H} \leftarrow [\frac{\partial}{\partial x} h({\mathbf{x}})_1 \vert_{{\mathbf{x_0}}}, \frac{\partial}{\partial x} h({\mathbf{x}})_2 \vert_{{\mathbf{x_0}}} \ldots \frac{\partial}{\partial x} h({\mathbf{x}})_{N_{inlier}} \vert_{{\mathbf{x_0}}}]$}
	\STATE{$d\mathbf{y} \leftarrow [ d\mathbf{y}_1 \; d\mathbf{y}_2 \ldots   d\mathbf{y}_{N_{inlier}}]$}
	\STATE{$d\hat{\mathbf{x}} = (\mathbf{H}^T\mathbf{W}\mathbf{H})^{-1}\mathbf{H}^T\mathbf{W}d\mathbf{y} $}
	\STATE{$       \mathbf{ \epsilon} = d\mathbf{y} - \mathbf{H}d\hat{\mathbf{x}} = (\mathbf{I} - \mathbf{H}(\mathbf{H}^T\mathbf{W}\mathbf{H})^{-1}\mathbf{H}^T\mathbf{W})d\mathbf{y} $}
	\STATE{$\lambda \leftarrow \epsilon^T\mathbf{W}\epsilon$}
	\STATE{$\delta \leftarrow \chi^2_{(3N-6,1-P_{fa})}$}
	\ENDWHILE

	\end{algorithmic}
\end{algorithm} 

\subsection{Protection Level Calculation}\label{PL}
In visual localization, we define protection level as the maximum translational error in each direction that the outlier detection algorithm is not expected to detect, denoted as $PL_x$, $PL_y$, $PL_z$, and they have to take account of both errors introduced by possible outliers and noise. A protection level for the total distance can also be calculated using the same approach, but we use axis-specific protection levels so that we can compare with $3\sigma$ method. We follow and modify the approach proposed by Walter~\cite{walter1995weighted} and Angus~\cite{angus2006raim} to calculate the protection level. Once the measurements pass the outlier rejection algorithm, we assume that there will be at most one feature outlier in the measurements. It is unlikely for the measurements to pass the outlier rejection test if there is more than 1 faulty feature in the measurements. In addition, depending on the application, we can always take account of a higher number of faulty features to obtain a more conservative protection level.

To analytically determine $PL_x$, $PL_y$, $PL_z$, we follow the problem formulation of~\cite{angus2006raim}. Because each feature produces 3 measurements, we denote the fault vector $\textbf{f}$ in Eq.~\ref{eq:fault} as
\begin{ceqn}
    \begin{align}
        \mathbf{f} = \mathbf{P}_j\mathbf{f}^*
    \end{align}
\end{ceqn}
where $\mathbf{f}^* \in \mathbb{R}^{3}$ , and $\mathbf{P}_j \in \mathbb{R}^{3N\times 3}$ in which each $3 \times 3$ matrix represents a feature. For $\mathbf{P}_j$, the $j$th $3 \times 3$ matrix is an identity matrix, and the rest of the $3 \times 3$ matrices are matrices of zeros. For example, if the first feature is an outlier and the rest of the features are inliers, then $\mathbf{P}_j$, is defined as  
 
\begin{align}
     \mathbf{P}_j = \mathbf{P}_1 = \begin{bmatrix} 1 & 0 & 0 \\ 0 & 1 & 0 \\ 0 & 0 & 1 \\ 0 & 0 & 0 \\ \vdots &\vdots& \vdots \\ 0 & 0 & 0\end{bmatrix}.
\end{align}
The weighted sum of squares residual $\lambda_f$ that is introduced by the fault vector $\mathbf{f}$ can be calculated as:

\begin{ceqn}
    \begin{align}
        \lambda_f =\mathbf{f}^T\mathbf{S}\mathbf{f} =  \mathbf{f^*}^T\mathbf{P}^T\mathbf{S}\mathbf{P}\mathbf{f}^*
    \end{align}
\end{ceqn}
where 
\begin{ceqn}
    \begin{align}
        \mathbf{S} =   \mathbf{W}(\mathbf{I} - \mathbf{H}(\mathbf{H}^T\mathbf{W}\mathbf{H})^{-1}\mathbf{H}^T\mathbf{W}).
    \end{align}
\end{ceqn}
Based on Eq.\ref{sol_linear}, the square of the outlier-induced error $(\varepsilon_{f})^2$ in the localization solution introduced by the fault vector $\mathbf{f}$ can also be calculated as:
\begin{ceqn}
    \begin{align}\label{eq:16}
        (\varepsilon_{f})^2 = \mathbf{f}^T\mathbf{D}_i\mathbf{f} = \mathbf{f^*}^T\mathbf{P}^T\mathbf{D}_i\mathbf{P}\mathbf{f}^*
    \end{align}
\end{ceqn}
where $\mathbf{D}_i$ depends on the direction and can be calculated as follows assuming the first three components of the pose correspond to the position:

\begin{ceqn}
    \begin{align}
        \mathbf{D}_i = \mathbf{W}\mathbf{H}(\mathbf{H}^T\mathbf{W}\mathbf{H})^{-1}\mathbf{A}^T_{i}\mathbf{A}_{i}(\mathbf{H}^T\mathbf{W}\mathbf{H})^{-1}\mathbf{H}^T\mathbf{W}
    \end{align}
\end{ceqn}
\begin{ceqn}
    \begin{align}
       \mathbf{A}_1 &= \begin{bmatrix} 1 & 0 & 0 & 0&0&0\end{bmatrix} \\
       \mathbf{A}_2 &= \begin{bmatrix} 0 & 1 & 0 & 0&0&0\end{bmatrix}\\ 
       \mathbf{A}_3 &= \begin{bmatrix} 0 & 0 & 1 & 0&0&0\end{bmatrix}.
    \end{align}
\end{ceqn}
where $i$ = 1, 2, 3 for $x$-direction, $y$-direction, and $z$-direction, respectively.
From Eq.~\ref{eq:16}, we find a larger vector $\mathbf{f^*}$ introduces a larger error in the localization solution. However, because the outlier rejection test is passed, the maximum value of $\mathbf{f^*}^T\mathbf{P}^T\mathbf{S}\mathbf{P}\mathbf{f}^*$ is constrained to be less than $\delta$. As a result, the vector $\mathbf{f}^*$ introduced by the outlier needs to allocated to maximize the outlier-induced error in the solution while still meeting the condition that $\mathbf{f^*}^T\mathbf{P}^T\mathbf{S}\mathbf{P}\mathbf{f}^* = \delta$~\cite{angus2006raim}. The square of the outlier-induced error that the protection level takes account of can be determined by solving the optimization problem:

\begin{equation}
    \begin{array}{rrclcl}
        \displaystyle  \max_{\mathbf{f}^*, \mathbf{P}_j \in \mathbf{P}} & \multicolumn{3}{l}{\mathbf{f^*}^T\mathbf{P}_j^T\mathbf{D}_i\mathbf{P}_j\mathbf{f}^*}\\
        \textrm{s.t.} & \mathbf{f^*}^T\mathbf{P}_j^T\mathbf{S}\mathbf{P}_j\mathbf{f}^* = \delta.
    \end{array}
\end{equation}

According to~\cite{angus2006raim}, we can eliminate $\mathbf{f}^*$ from the equation and simplify this constrained optimization problem to an unconstrained optimization problem as follows:
\begin{equation}
   \begin{array}{rrclcl}\label{eq:simplified op}
        \displaystyle  \max_{\mathbf{P}_j \in \mathbf{P}} & \multicolumn{3}{l}{	\mathbf{\Lambda}_{\max}(\mathbf{D}_i,\mathbf{P}_j,\mathbf{S})\delta}
    \end{array}
\end{equation}
where $\mathbf{\Lambda}_{\max}(\mathbf{D}_i,\mathbf{P}_j,\mathbf{S})$ is the largest eigenvalue of 
\begin{ceqn}
    \begin{align}
        \mathbf{P}_j^T\mathbf{D}_i\mathbf{P}_j(\mathbf{P}_j^T\mathbf{S}\mathbf{P}_j)^{-1}.
    \end{align}
\end{ceqn}
We can iterate through all the features to find the $\mathbf{P}_j$ that gives the maximum value of Eq.~\ref{eq:simplified op}, which is defined as: 
\begin{equation}
    \mathbf{P}^* = \begin{array}{rrclcl}
        \displaystyle  \argmax_{\mathbf{P}_j \in \mathbf{P}} & \multicolumn{3}{l}{	\mathbf{\Lambda}_{\max}(\mathbf{D}_i,\mathbf{P}_j,\mathbf{S})\delta}.
    \end{array}
\end{equation}
As a result, the component of the protection level that takes account of the outlier-induced error can be found as:

\begin{ceqn}
    \begin{align}
\varepsilon_f &= \sqrt{	\mathbf{\Lambda}_{\max}(\mathbf{D}_i,\mathbf{P}^*,\mathbf{S})\delta}.
    \end{align}
\end{ceqn}
The noise-induced error, $\varepsilon_n$, in the localization solution can be calculated using the following equations:
\begin{ceqn}
    \begin{align}\label{eq:pl_noise}
       \varepsilon_n &=  k\sqrt{[(\mathbf{H}^T\mathbf{W}\mathbf{H})^{-1}]_{i,i}}
    \end{align}
\end{ceqn}
where $i$ = 1, 2, 3 for $x$-direction, $y$-direction, $z$-direction, respectively, and $k$ is the number of standard deviations corresponding to the specified detection probability $P_d$. In real-life applications, $P_d$ is normally set to be 99.73\% and the corresponding $k = 3$. Incorporating both the outlier-induced error and the noise-induced error, we can define the protection level for visual localization as follows:
\begin{flalign} 
       \label{eq:plx}
        PL_i &= \varepsilon_f + \varepsilon_n \\ &=\sqrt{	\mathbf{\Lambda}_{\max}(\mathbf{D}_i,\mathbf{P}^*,\mathbf{S})\delta}+k\sqrt{[(\mathbf{H}^T\mathbf{W}\mathbf{H})^{-1}]_{i,i}}
\end{flalign}

where $i$ = 1, 2, 3 for $x$-direction, $y$-direction, and $z$-direction, respectively. $PL_x$, $PL_y$, $PL_z$ are equal to $PL_1$, $PL_2$, and $PL_3$, respectively.
\section{PERFORMANCE EVALUATION}\label{metric}
The best error bound for an estimation process should be as tight as possible while still bounding the error at all times. However, for nonlinear systems with outliers, it is not possible to guarantee that a proposed bound will successfully bound the true error at all times.  There are currently no existing error bound metrics that quantitatively evaluate potential error bound performance. We therefore propose a novel relaxed bound tightness (RBT) metric that quantitatively evaluates the performance of error bounds for applications where the error is assumed to follow a Gaussian distribution. The metric, $\mathcal{Z}$, is calculated as follows:  
\begin{ceqn}
    \begin{align}
       \mathcal{Z} = \sqrt{\frac{\sum_{i=1}^{N}\varrho
 (\frac{\nu_i -e_i}{\sigma_i})^2}{N}}
    \end{align}
\end{ceqn}
where $\nu_i$ and $e_i$ are respectively the error bound and the error for a sample $i$, $N$ is the number of samples, $\sigma_i$ is the covariance of the error for the given sample, and $\varrho$ is a weight function defined as follows:
\begin{equation}
\varrho = \begin{dcases} 1 & \nu_i \geq e_i \\
\tau &  \nu_i < e_i 
\end{dcases}.
\end{equation}
Because the error is supposed to be an upper bound rather than a prediction of the expectation of the errors, the weight function $\varrho$ should penalize error bounding failures more heavily than loose bounds.  The coefficient $\tau$ decides to what extent the metric favors a tight error bound with a higher probability of failure over a conservative bound with a low probability of failure and vice versa, and is set to depend on a user-defined detection probability $P_d$. This probability $P_d$ is the minimum probability that the method correctly bounds the error required by the user. 

To determine the value of $\tau$, we assume the error follows a Gaussian distribution without outliers.  Let $\Phi^{-1}$ be the quantile function for a Gaussian distribution, and define the ideal error bound, $\upsilon^*$, to be 
\begin{equation}
\upsilon^* = \Phi^{-1}\left(1-\frac{1-P_d}{2}\right).
\end{equation} 
This choice of bound corresponds exactly with the definition of the detection probability $P_d$ for a Gaussian distribution. Given the bound, $\upsilon^*$, we now solve for the value of $\tau$ that minimizes the metric $\mathcal{Z}$. Note that in the RBT metric, because we divide the difference of the bound and the true errors by the covariance for each sample, the value of $\tau$ is independent of the value of $\sigma_i$ given that  the error is assumed to follow a Gaussian distribution.

For a safety-critical application, $P_d$ should be defined close to 100\% such that this metric prefers a method that is conservative, but is able to correctly bound the error for a higher percentage of time. For applications where safety is less critical, $P_d$ can be defined less aggressively, such that the metric favors a method that provides tighter bounds but might fail to bound the error at certain times. 

\section{EXPERIMENTAL RESULTS}\label{results}
In order to validate the proposed method and determine if the protection level correctly bounds the translational errors in each direction, we perform experiments on the machine hall sequences in the well-known EuRoC dataset~\cite{burri2016euroc}. The EuRoC dataset is collected using a micro aerial vehicle (MAV), and the ground truth poses of the MAV in the machine hall are captured by a Leica MS50 laser tracker. 

We take advantage of the localization mode of ORB-SLAM2~\cite{mur2017orb} and add the proposed outlier rejection and protection level calculation modules. We first construct the map using the ground truth poses and then run the localization mode of ORB-SLAM2 with the proposed integrity monitoring algorithm to obtain the estimated poses and the protection levels for each frame. For stereo vision-based localization, no initialization is needed to recover the scale. As a result, we start the estimation when the MAV is static after the initialization phase at the start of each sequence. Afterwards, we align the ground truth with the estimated trajectory and evaluate the error. Finally, we compare the protection level and $3\sigma$ bound in each direction with the true translational error for each frame. 

\begin{figure}[ht] 
	\begin{center}
		\includegraphics[width=0.9\linewidth]{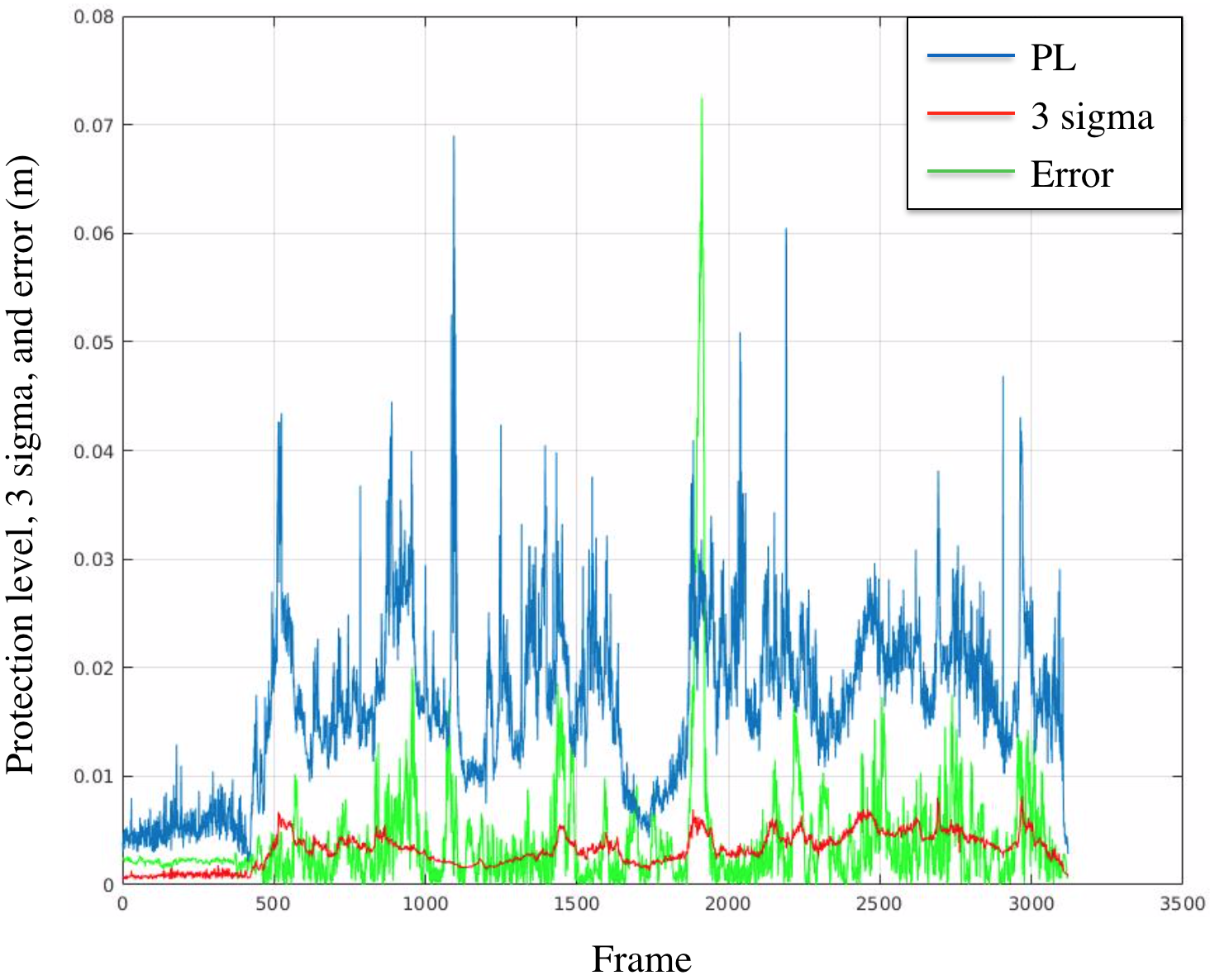}
	\end{center}
		\caption{Protection Level, $3\sigma$ and errors vs. frames  in $z$-direction for the \textit{MH\_easy\_01} sequence in EuRoc dataset using $1$-pixel covariance assumption. The $3\sigma$ approach only correctly bounds the error around 54\%, and the protection level correctly bounds the error for more than 95\%.}
		\label{fig:mh01_z}
\end{figure}

In this experiment, the false alert probability $P_{fa}$ is set to be 0.05 and $k$ in Eq.~\ref{eq:pl_noise} is set to be 3, which are the typical values used for robotics applications~\cite{ahn2012board}. In ORB-SLAM2, the covariance matrix $\mathbf{Q}_j$ for each feature is related to the scale at which the keypoint is extracted~\cite{mur2017orb}. Because ORB-SLAM2 is a vision only SLAM algorithm, it is the relative value between the covariances of the features that decides their importance in determining the pose of the camera rather than the absolute value of the covariances. At first, we perform the experiments using the default ORB-SLAM2 covariance value, which is set to be 1 pixel for features detected in the base scale level. Fig.~\ref{fig:mh01_z} shows the protection level, $3\sigma$ bound, and the true errors in the $z$-direction for the entire trajectory of \textit{MH\_01\_easy} using the 1-pixel assumption. It is found that the protection level is able to bound the error for more than 95\% of the frames but the $3\sigma$ is only able to bound around 50\% of the frames. Possible causes for this drastic failure of the $3\sigma$ bound are first that there are outlier features in these frames where the $3\sigma$ does not bound the error and second that the covariance assumed for the inlier measurement noise on the measurement model is too small. To avoid the second case, we perform the experiments with the assumptions of 1, 1.5 and 2 pixel covariance for features detected in the base scale level. These are the typical values used for the covariance of the camera measurement model~\cite{huang2014towards, wu2015square}. 

\begin{figure*}%
\centering
\begin{subfigure}{0.95\columnwidth}
\includegraphics[width=\columnwidth]{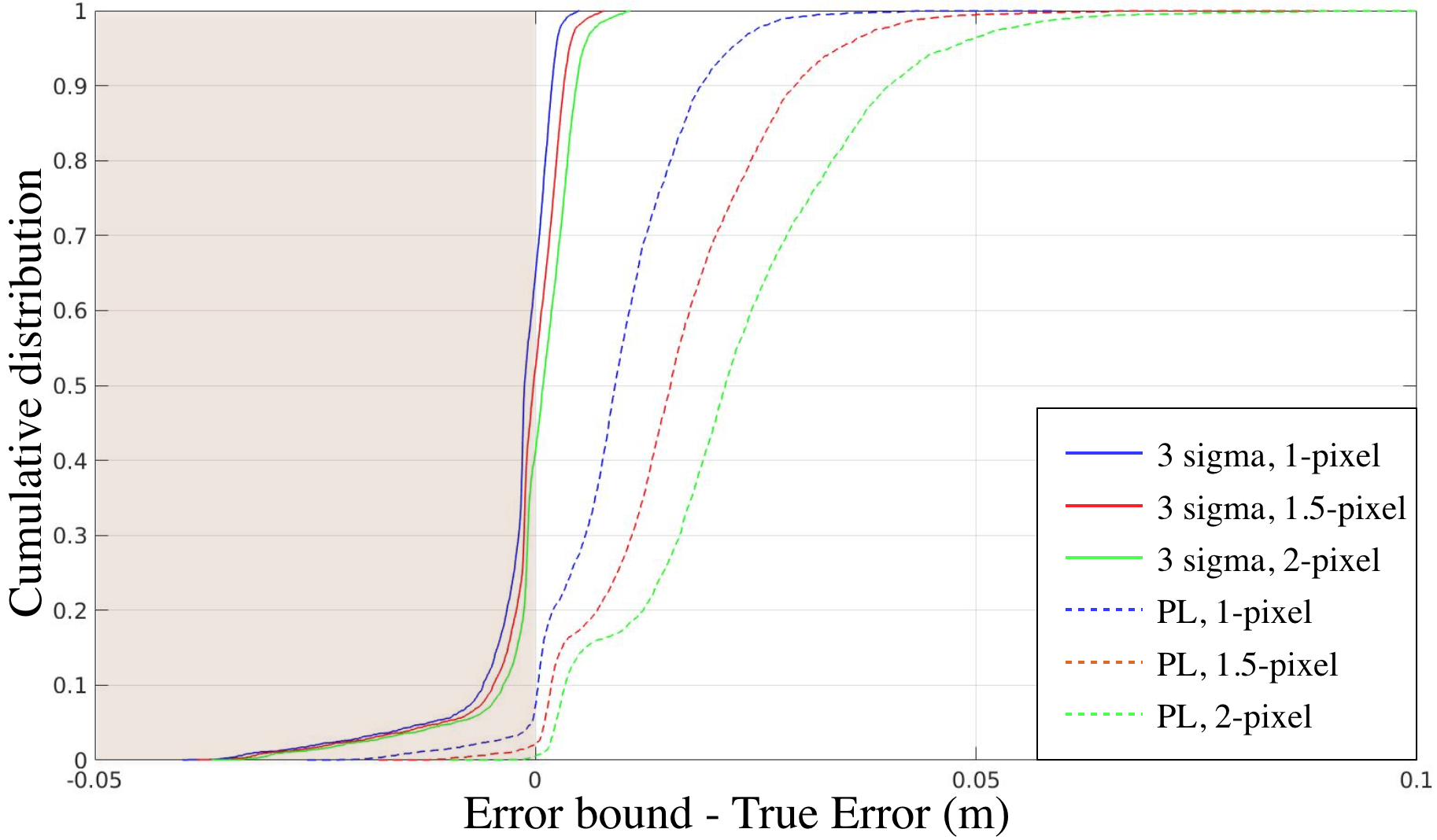}%
\caption{MH\_01\_easy}%
\end{subfigure}\hfill%
\begin{subfigure}{0.95\columnwidth}
\includegraphics[width=\columnwidth]{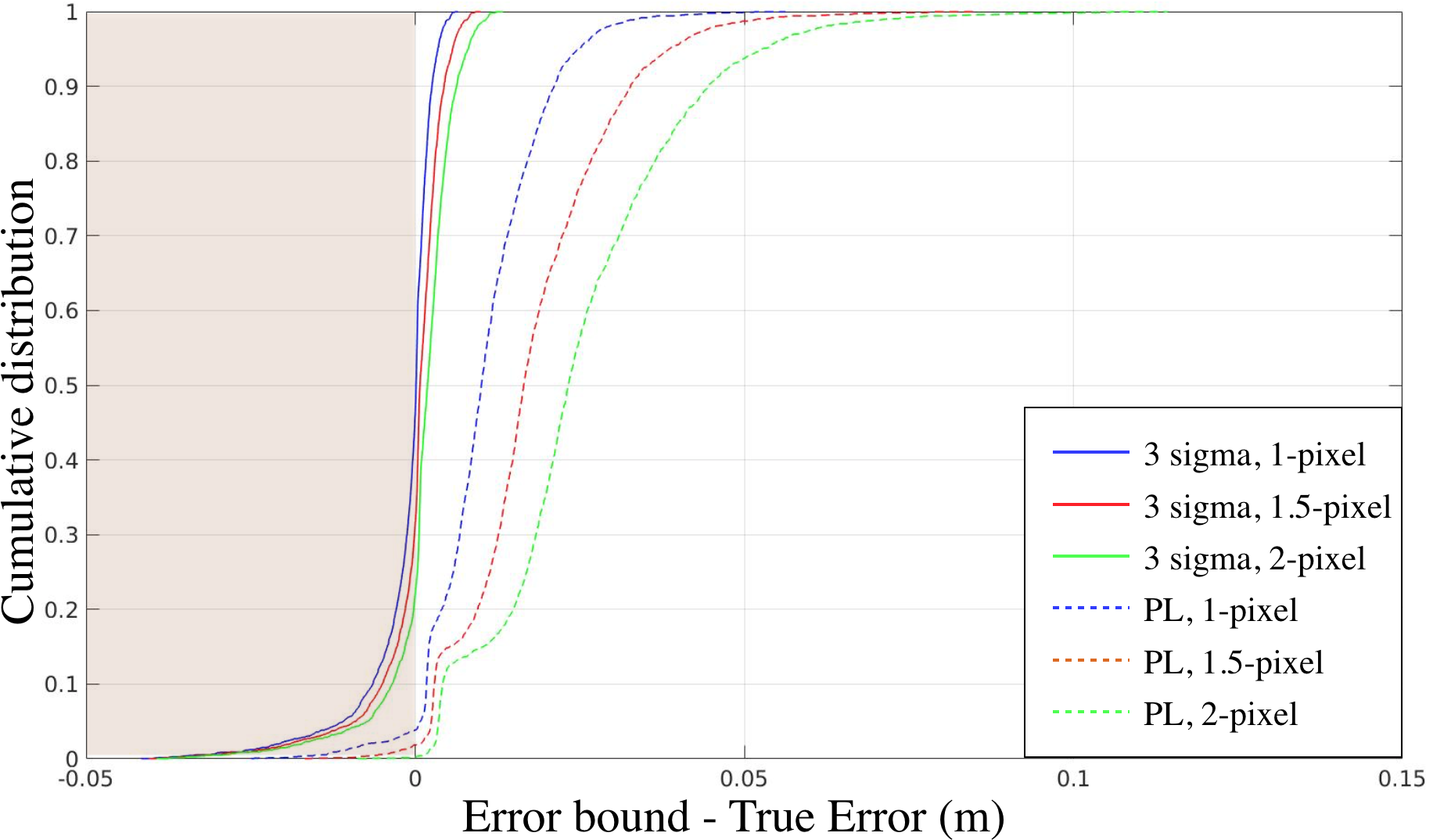}%
\caption{MH\_02\_easy}%
\end{subfigure}\hfill%
\begin{subfigure}{0.95\columnwidth}
\includegraphics[width=\columnwidth]{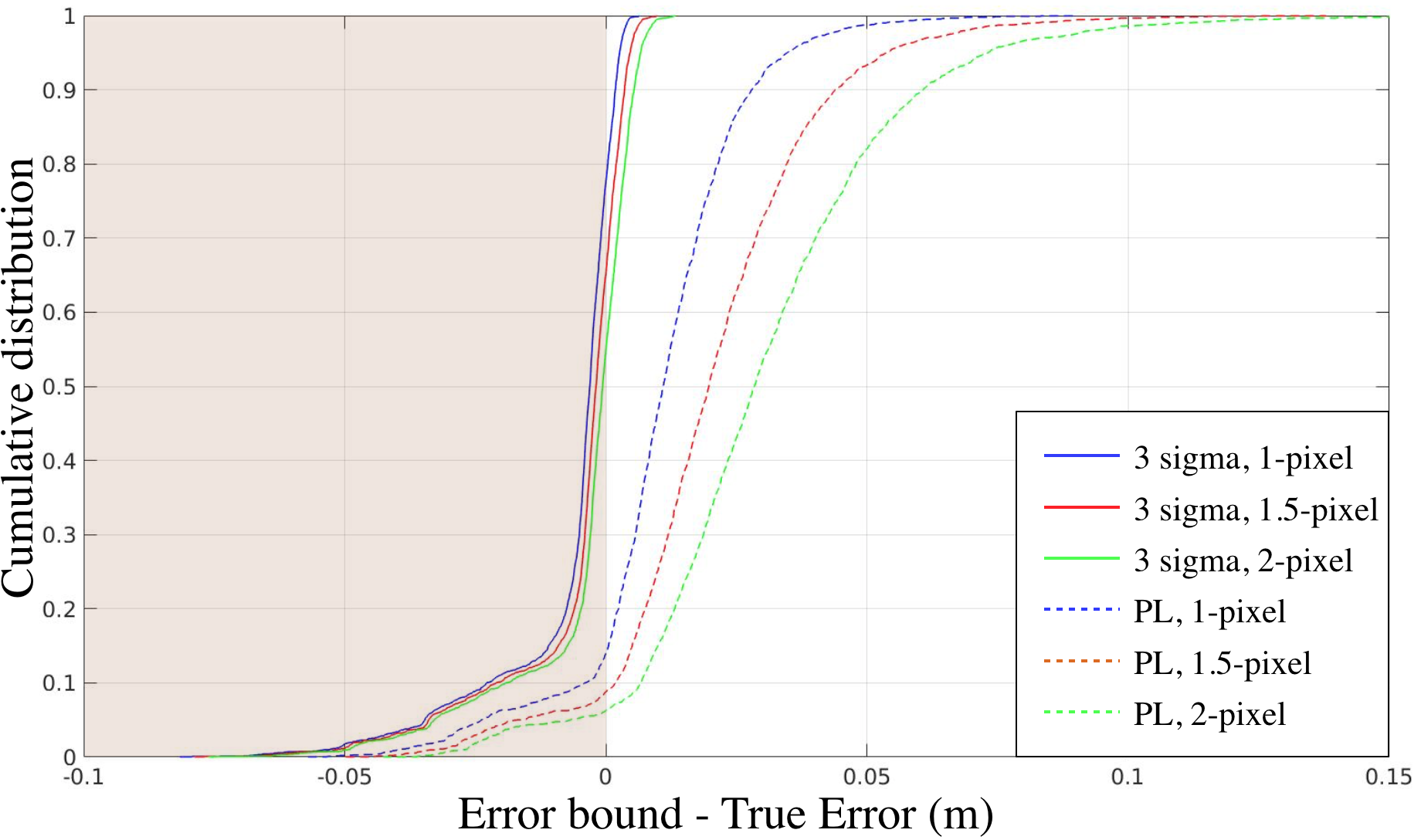}%
\caption{MH\_03\_medium}%
\end{subfigure}\hfill%
\begin{subfigure}{0.95\columnwidth}
\includegraphics[width=\columnwidth]{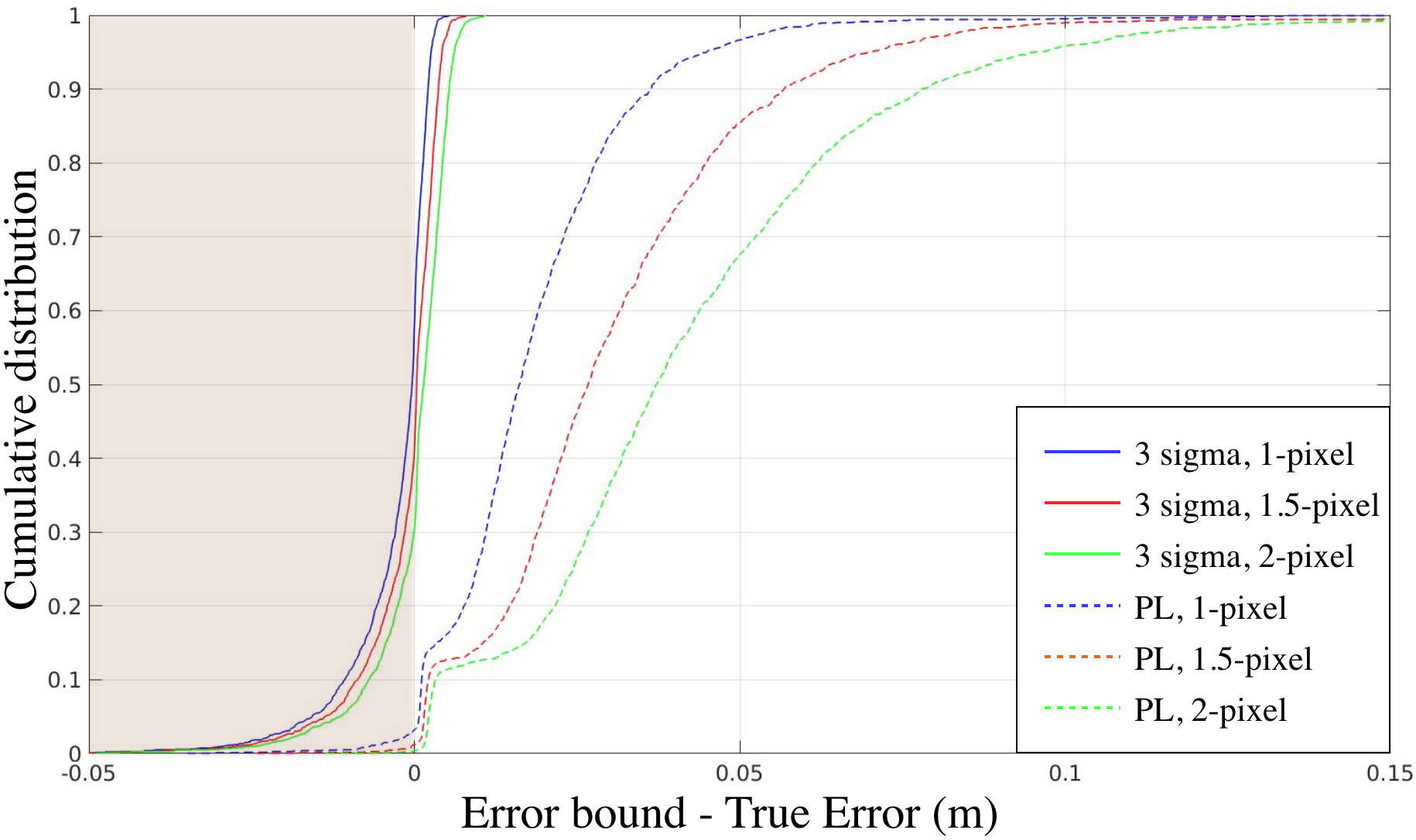}%
\caption{MH\_04\_difficult}%
\end{subfigure}\hfill%
\begin{subfigure}{0.95\columnwidth}
\includegraphics[width=\columnwidth]{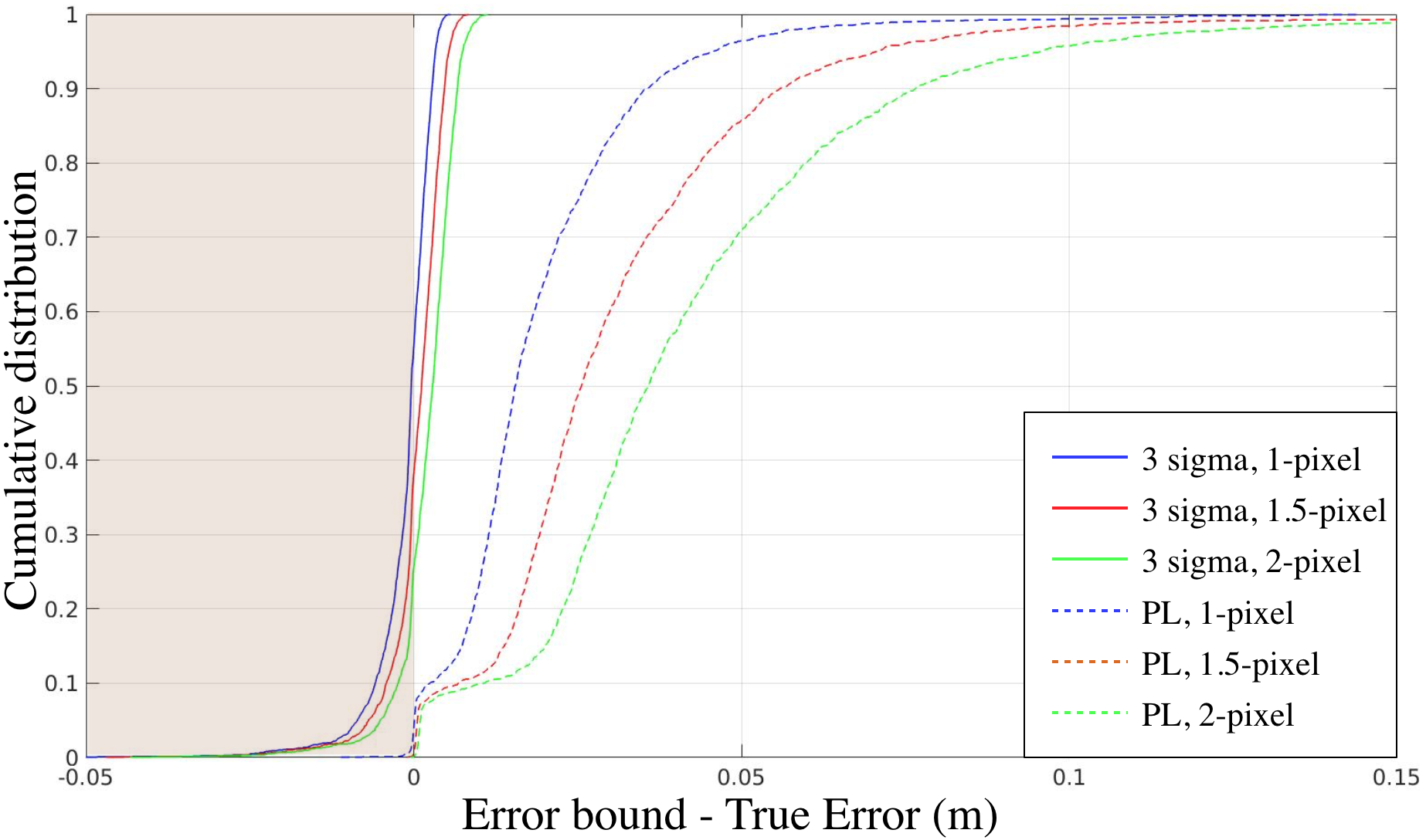}%
\caption{MH\_05\_difficult}%
\end{subfigure}\hfill%
\caption{Cumulative distribution plots for the difference between the error bound and the error for the machine hall sequences in the EuRoc dataset. Compared with $3\sigma$ method, the protection level correctly bounds the error for a higher percentage of frames for all sequences.}
\label{fig:cdf}
\end{figure*}

In Fig.~\ref{fig:cdf}, we compare the protection level method and the $3\sigma$ method by plotting the cumulative distribution of the difference between the error bound and the true error in $x$-direction for different assumptions of the measurement covariance values. The ideal error bound would be entirely contained on the positive side of the x-axis in the plot, because we want the error bound to bound the true error exclusively, and also distributed as close to 0 as possible, because we want the error bound to be as tight as possible. The intersection between the curves and the vertical line $x = 0$ gives the percentage of frames that the error bound method fails to bound the error. As expected, the protection level provides a more conservative error bound compared with $3\sigma$ but is able to bound the error for a higher percentage of frames. If we use the 1-pixel assumption, which is the default covariance value of ORB-SLAM2, we find that $3\sigma$ is only able to correctly bound the error in $x$-direction for about 35\% of the frames for \textit{MH\_01\_easy}, 54\% for \textit{MH\_02\_easy}, 22\% for \textit{MH\_03\_medium}, 43\% for \textit{MH\_04\_difficult}, and 45\% for \textit{MH\_05\_difficult}. However, the protection level is able to bound the error about 85\% for \textit{MH\_03\_medium}, and around 95\% for the rest of the sequences. If we use 1.5-pixel and 2-pixel assumptions, the $3\sigma$ method is able to correctly bound at a higher percentage rate but still significantly less than the protection level method, which correctly bounds the error at approximately 95\% for \textit{MH\_03\_medium} and near 100\% for the rest of the sequences. Similar results hold for $y$-direction and $z$-direction. 

We also compare the proposed protection level with the $3\sigma$ method using the metric proposed in Sec.~\ref{metric}. We set the parameter $P_d$ to be 99.73\%, because this is one of the most commonly used detection probabilities in real-life applications and also corresponds with the $3\sigma$ bound used for comparison. Since we determine the penalty coefficient $\tau$ in the metric using Gaussian distribution and choose $P_d$ to be 99.73\%, the metric favors $3\sigma$ method if the error truly follows a zero-mean Gaussian distribution. However, from Table~\ref{performance}, it is found that the proposed protection level outperforms the $3\sigma$ method for all sequences for all 3 covariance assumptions. The results show that feature-based visual measurements are susceptible to error, and thus the error in the position solution does not perfectly follow a Gaussian distribution. Further, the protection level approach is able to correctly bound the error at a higher percentage rate and outperform $3\sigma$ on the chosen metric. 

\captionsetup[table]{skip=0pt}
\begin{table*}[!ht] 
\centering 
    \captionof{table}{Performance comparison using the proposed RBT metric}
    \label{performance}
 \begin{tabular}{||c | c| c c| c c| c c||} 
 \hline
 \multirow{2}{*}{Sequence} &  \multirow{2}{*}{Axis} & \multicolumn{2}{c|}{1 pixel} &\multicolumn{2}{c|}{1.5 pixels}& \multicolumn{2}{c||}{2 pixels} \\
     &  & PL & $3\sigma$ & PL & $3\sigma$ & PL & $3\sigma$ \\ 
 \hline
\multirow{3}{*}{MH\_01\_easy} &x & 126.2 & 398.0 & 38.1 & 235.1 & \textbf{17.4} & 159.3 \\ 
                              &y & 59.8  & 358.3 & \textbf{18.8} & 203.1 & 19.0 & 134.5 \\ 
                              &z & 127.3 & 206.0 & 45.7 & 104.6 & \textbf{25.7} & 68.6  \\ \hline
\multirow{3}{*}{MH\_02\_easy} &x & 121.2 & 317.3  &30.8 & 179.1 & \textbf{20.3} & 103.8 \\ 
                              &y & 60.5  & 353.9  &25.2 & 205.7 & \textbf{21.7} & 134.8 \\ 
                              &z & 95.2  & 221.2  &25.7 & 125.0 & \textbf{23.7} & 71.1  \\  \hline
\multirow{3}{*}{MH\_03\_medium} &x & 1053.5 & 1277.5 & 641.1 & 830.1 & \textbf{441.0} & 607   \\
                                &y & 682.2  & 971.8  & 389.3 & 622.7 & \textbf{252.3} & 449.8 \\
                                &z & 1056.1 & 1265.3 & 609.1 & 819.9 & \textbf{397.0} & 598.5 \\  \hline
\multirow{3}{*}{MH\_04\_difficult} &x & 143.8 & 340.9 & 48.1 & 200.9 &\textbf{21.8}  & 132.6 \\ 
                                   &y & 105.3 & 336.6 & 41.9 & 199.5 & \textbf{26.3} & 133.4\\ 
                                   &z &  89.8 & 165.5 & 48.7 & 88.7  & \textbf{32.2} & 53.8 \\ \hline
\multirow{3}{*}{MH\_05\_difficult} &x & 51.8 & 256.6 & 25.6 & 146.9 & \textbf{21.8} &132.6\\ 
                                   &y & 30.0 & 215.3 & 28.3 & 118.1 & \textbf{26.3} &133.4\\ 
                                   &z & 50.1 & 173.3 & \textbf{26.9} & 93.8  & 32.2 &53.8 \\  \hline
\end{tabular}
\end{table*}

The results for both protection level and $3\sigma$ are significantly worse for the \textit{MH\_03\_medium} sequence, because the true errors obtained in this sequence are significantly larger than other sequences. The hypothesis is that the map points might contain a number of errors for this sequence. To verify this hypothesis, we perform the experiments using the map constructed by the SLAM mode of ORB-SLAM2, instead of the map constructed using ground truth, and evaluate the localization solution against the SLAM solution instead of the ground truth poses, which removes the effect of the map errors on the solution. For the \textit{MH\_03\_medium} sequence, it is found that the translation RSME is 0.011 m in this case which is much smaller than 0.032 m which is the RSME obtained if we evaluate the errors against the ground truth. The performance for protection level in the $x$-direction evaluated using the proposed metric is 17.8, 18.4, and 18.7 for the 1-pixel, 1.5-pixel, and 2-pixel assumptions, respectively, while the performance for the $3\sigma$ approach is 90.0, 43.2, and 23.2. The protection level still outperforms the $3\sigma$ method using the proposed RBT metric. The same holds for the other two directions and other sequences. 

We notice that although protection level is able to bound the error better, it produces a looser error bound compared with $3\sigma$. The potential issue is that it could result in more false positive warnings during the operation. We believe the selection of the error bound should depend on the safety standard of the application, and the proposed protection level is more suitable for life-critical applications. 

\section{CONCLUSIONS AND FUTURE WORK}
This work presents an integrity monitoring algorithm to estimate the maximum possible translational error in the visual localization solution. The framework is inspired by RAIM and modified to fit the problem formulation of visual localization. It first detects outliers based on the Parity Space Method and then calculates the maximum possible error the outlier detection algorithm is not expected to detect. In addition, we proposed a relaxed bound tightness metric to quantitatively evaluate the performance of error bounds. Finally, by performing experiments on the EuRoC dataset and evaluating the results using the proposed metric, it is determined that the proposed protection level produces more reliable bounds than the typical $3\sigma$ method and provides an approach to assess the integrity of the solution. Future work will include taking the uncertainty of the map and possible incorrect feature covariance into considerations when calculating protection level, and also further developing the concept to provide integrity monitoring for visual odometry.

{
\bibliographystyle{IEEEtran}
\bibliography{IEEEtran}

\begin{thebibliography}{10}
\providecommand{\url}[1]{#1}
\csname url@rmstyle\endcsname
\providecommand{\newblock}{\relax}
\providecommand{\bibinfo}[2]{#2}
\providecommand\BIBentrySTDinterwordspacing{\spaceskip=0pt\relax}
\providecommand\BIBentryALTinterwordstretchfactor{4}
\providecommand\BIBentryALTinterwordspacing{\spaceskip=\fontdimen2\font plus
\BIBentryALTinterwordstretchfactor\fontdimen3\font minus
  \fontdimen4\font\relax}
\providecommand\BIBforeignlanguage[2]{{%
\expandafter\ifx\csname l@#1\endcsname\relax
\typeout{** WARNING: IEEEtran.bst: No hyphenation pattern has been}%
\typeout{** loaded for the language `#1'. Using the pattern for}%
\typeout{** the default language instead.}%
\else
\language=\csname l@#1\endcsname
\fi
#2}}

\bibitem{mur2017orb}
R.~Mur-Artal and J.~D. Tard{\'o}s, ``Orb-slam2: An open-source slam system for
  monocular, stereo, and rgb-d cameras,'' \emph{IEEE Transactions on Robotics},
  vol.~33, no.~5, pp. 1255--1262, 2017.

\bibitem{leutenegger2013keyframe}
S.~Leutenegger, P.~Furgale, V.~Rabaud, M.~Chli, K.~Konolige, and R.~Siegwart,
  ``Keyframe-based visual-inertial slam using nonlinear optimization,''
  \emph{Proceedings of Robotis Science and Systems (RSS) 2013}, 2013.

\bibitem{ochieng2002assessment}
W.~Ochieng, K.~Sheridan, K.~Sauer, X.~Han, P.~Cross, S.~Lannelongue, N.~Ammour,
  and K.~Petit, ``An assessment of the raim performance of a combined
  galileo/gps navigation system using the marginally detectable errors (mde)
  algorithm,'' \emph{GPS Solutions}, vol.~5, no.~3, pp. 42--51, 2002.

\bibitem{walter1995weighted}
T.~Walter and P.~Enge, ``Weighted raim for precision approach,'' in
  \emph{PROCEEDINGS OF ION GPS}, vol.~8.\hskip 1em plus 0.5em minus 0.4em\relax
  Institute of Navigation, 1995, pp. 1995--2004.

\bibitem{2019imekf}
G.~D. Arana, M.~Joerger, and M.~Spenko, ``Efficient integrity monitoring for
  kf-based localization,'' in \emph{2019 IEEE International Conference on
  Robotics and Automation (ICRA)}.\hskip 1em plus 0.5em minus 0.4em\relax IEEE,
  2019, pp. 6374--6380.

\bibitem{2019imekf2}
G.~D. Arana, O.~A. Hafez, M.~Joerger, and M.~Spenko, ``Recursive integrity
  monitoring for mobile robot localization safety,'' in \emph{2019 IEEE
  International Conference on Robotics and Automation (ICRA)}.\hskip 1em plus
  0.5em minus 0.4em\relax IEEE, 2019, pp. 305--311.

\bibitem{tong2011batch}
C.~H. Tong and T.~D. Barfoot, ``Batch heterogeneous outlier rejection for
  feature-poor slam,'' in \emph{2011 IEEE International Conference on Robotics
  and Automation}.\hskip 1em plus 0.5em minus 0.4em\relax IEEE, 2011, pp.
  2630--2637.

\bibitem{das2014outlier}
A.~Das and S.~L. Waslander, ``Outlier rejection for visual odometry using
  parity space methods,'' in \emph{2014 IEEE International Conference on
  Robotics and Automation (ICRA)}.\hskip 1em plus 0.5em minus 0.4em\relax IEEE,
  2014, pp. 3613--3618.

\bibitem{liu2005gps}
J.~Liu, M.~Lu, Z.~Feng, and J.~Wang, ``Gps raim: statistics based improvement
  on the calculation of threshold and horizontal protection radius,'' in
  \emph{International Symposium on GPS/GNSS}, 2005, pp. 8--10.

\bibitem{brown1994gps}
R.~G. Brown, \emph{GPS RAIM: Calculation of Thresholds and Protection Radius
  Using Chi-square Methods; a Geometric Approach}.\hskip 1em plus 0.5em minus
  0.4em\relax Radio Technical Commission for Aeronautics, 1994.

\bibitem{angus2006raim}
J.~Angus, ``Raim with multiple faults,'' \emph{Navigation}, vol.~53, no.~4, pp.
  249--257, 2006.

\bibitem{engel2014lsd}
J.~Engel, T.~Sch{\"o}ps, and D.~Cremers, ``Lsd-slam: Large-scale direct
  monocular slam,'' in \emph{European conference on computer vision}.\hskip 1em
  plus 0.5em minus 0.4em\relax Springer, 2014, pp. 834--849.

\bibitem{fischler1981random}
M.~A. Fischler and R.~C. Bolles, ``Random sample consensus: a paradigm for
  model fitting with applications to image analysis and automated
  cartography,'' \emph{Communications of the ACM}, vol.~24, no.~6, pp.
  381--395, 1981.

\bibitem{kitt2010visual}
B.~Kitt, A.~Geiger, and H.~Lategahn, ``Visual odometry based on stereo image
  sequences with ransac-based outlier rejection scheme,'' in \emph{2010 ieee
  intelligent vehicles symposium}.\hskip 1em plus 0.5em minus 0.4em\relax IEEE,
  2010, pp. 486--492.

\bibitem{scaramuzza20111}
D.~Scaramuzza, ``1-point-ransac structure from motion for vehicle-mounted
  cameras by exploiting non-holonomic constraints,'' \emph{International
  journal of computer vision}, vol.~95, no.~1, pp. 74--85, 2011.

\bibitem{bar2004estimation}
Y.~Bar-Shalom, X.~R. Li, and T.~Kirubarajan, \emph{Estimation with applications
  to tracking and navigation: theory algorithms and software}.\hskip 1em plus
  0.5em minus 0.4em\relax John Wiley \& Sons, 2004.

\bibitem{tzoumas2019outlier}
V.~Tzoumas, P.~Antonante, and L.~Carlone, ``Outlier-robust spatial perception:
  Hardness, general-purpose algorithms, and guarantees,'' \emph{arXiv preprint
  arXiv:1903.11683}, 2019.

\bibitem{barfoot2017state}
T.~D. Barfoot, \emph{State Estimation for Robotics}.\hskip 1em plus 0.5em minus
  0.4em\relax Cambridge University Press, 2017.

\bibitem{burri2016euroc}
M.~Burri, J.~Nikolic, P.~Gohl, T.~Schneider, J.~Rehder, S.~Omari, M.~W.
  Achtelik, and R.~Siegwart, ``The euroc micro aerial vehicle datasets,''
  \emph{The International Journal of Robotics Research}, vol.~35, no.~10, pp.
  1157--1163, 2016.

\bibitem{ahn2012board}
S.~Ahn, S.~Yoon, S.~Hyung, N.~Kwak, and K.~S. Roh, ``On-board odometry
  estimation for 3d vision-based slam of humanoid robot,'' in \emph{2012
  IEEE/RSJ International Conference on Intelligent Robots and Systems}.\hskip
  1em plus 0.5em minus 0.4em\relax IEEE, 2012, pp. 4006--4012.

\bibitem{huang2014towards}
G.~Huang, M.~Kaess, and J.~J. Leonard, ``Towards consistent visual-inertial
  navigation,'' in \emph{2014 IEEE International Conference on Robotics and
  Automation (ICRA)}.\hskip 1em plus 0.5em minus 0.4em\relax IEEE, 2014, pp.
  4926--4933.

\bibitem{wu2015square}
K.~Wu, A.~Ahmed, G.~A. Georgiou, and S.~I. Roumeliotis, ``A square root inverse
  filter for efficient vision-aided inertial navigation on mobile devices.'' in
  \emph{Robotics: Science and Systems}, vol.~2, 2015.

\end{thebibliography}
}

\end{document}